\documentclass[conference]{IEEEtran}
\IEEEoverridecommandlockouts
\usepackage{cite}
\usepackage{amsmath,amssymb,amsfonts}
\usepackage{algorithm, algpseudocode}
\usepackage{graphicx, subcaption}
\usepackage{textcomp}
\usepackage{xcolor}
\def\BibTeX{{\rm B\kern-.05em{\sc i\kern-.025em b}\kern-.08em
    T\kern-.1667em\lower.7ex\hbox{E}\kern-.125emX}}
\begin{document}

\title{Rotation-Based Subspace Tracking for Robust Kernel PCA on Streaming Data\\
\thanks{
\IEEEauthorrefmark{1}Corresponding author
\IEEEauthorrefmark{2}Majority of research performed while also affiliated with George Washington University, Washington, DC, USA}
}

\author{
\IEEEauthorblockN{
Kris Lokere\IEEEauthorrefmark{1}
}
\IEEEauthorblockA{
\textit{Harvard University}, Cambridge, MA, USA\IEEEauthorrefmark{2} \\
klokere@g.harvard.edu\\
}
\and
\IEEEauthorblockN{
John Fossaceca
}
\IEEEauthorblockA{
\textit{George Washington University}, Washington, DC, USA \\
jfossaceca@email.gwu.edu\\
}
}

\maketitle

\begin{abstract}
Machine learning models process large amounts of data, and Principal Component Analysis (PCA) is a widely used technique to reduce the dimensionality of the data and extract useful features. In practice, datasets often change over time (data drift) and/or arrive one sample at a time (streaming data), making it infeasible to process the entire dataset at once in batch mode. Real-world data also often contains nonlinear patterns, which traditional PCA cannot extract. Kernel PCA addresses this by implicitly mapping samples into a Reproducing Kernel Hilbert Space (RKHS). Raw data also often contains outliers, which can have an outsized effect on the estimated subspace unless the algorithm is made robust. However, existing online robust kernel PCA algorithms are designed to converge to a subspace that is assumed to be fixed, and gradient-descent-based updates lose their effectiveness at tracking further changes once this initial alignment is achieved. This paper introduces a rotation-based update mechanism, which updates the subspace estimate by rotating it toward each new incoming feature vector in Reproducing Kernel Hilbert Space, rather than relying on gradient descent alone. We present two complementary rotation strategies, and show that the extent of rotation can be moderated by a robust influence function to mitigate the effect of outliers. Through experiments on synthetic streaming data with a known ground-truth subspace, we show that per-sample rotations converge faster than gradient descent alone, demonstrating an effective mechanism for dynamically tracking a nonlinear subspace in streaming data.
\end{abstract}


\section{Introduction}
Machine learning models are trained on large amounts of data. When the number of features in the data is very high, processing cost becomes expensive, interpretability suffers, and performance of algorithms can deteriorate because of the "curse of dimensionality." Principal Component Analysis (PCA) is a technique to reduce the dimensionality of the data, in a way that lowers processing cost, minimizes information loss, and improves interpretability~\cite{jolliffe2016}.

In practice, datasets often change over time (data drift) and/or arrive one sample at a time (streaming data). This makes it infeasible to process the entire dataset at once in batch mode. An improvement to batch processing is to calculate the PCA in an online manner, updating the calculation each time that new data is observed, without needing to recompute everything from scratch~\cite{dasgupta2026}. Real-world data also often contains nonlinear patterns, which traditional PCA cannot extract. Kernel PCA has been developed to extract nonlinear features from such data by implicitly mapping samples into a Reproducing Kernel Hilbert Space (RKHS) \cite{scholkopf1997}, \cite{tonin2024}.

While online robust kernel PCA algorithms exist, many are designed under the assumption that the underlying distribution is drawn from a static, unchanging subspace, and the focus is on getting the estimate to converge to that fixed subspace. This assumption breaks down for data that drifts over time, and a time-dependent estimation of the subspace is required instead. Existing iterative robust kernel PCA algorithms that update their subspace estimate via gradient descent, such as the method of Huang and Yeh \cite{huang2011}, are provably convergent to a batch solution under mild assumptions, but this convergence guarantee applies only when the true subspace is fixed. Once initial convergence is achieved, gradient-based updates lose much of their effectiveness at tracking further changes in the underlying subspace, because the projection of new data onto the current subspace estimate leaves little residual signal to drive further movement.

This paper addresses this limitation by introducing a rotation-based update mechanism for online kernel PCA. Rather than relying solely on gradient descent, the estimated subspace is rotated—using a per-sample rotation matrix constructed in RKHS—toward the direction of each new incoming feature vector. This rotation is an original contribution: to the best of our knowledge, no prior online kernel PCA algorithm uses rotation matrices as an update mechanism for dynamically tracking a nonlinear subspace. We introduce two complementary rotation strategies, and show how the extent of rotation applied at each step can be moderated by a robust influence function, so that the effect of outliers on the estimated subspace remains bounded.

We validate this mechanism through controlled experiments on synthetic streaming data with a known ground-truth subspace. We show that enabling the rotation-based update leads to measurably faster convergence than relying on gradient descent alone, and that the mechanism naturally aligns individual basis vectors of the estimated subspace with the corresponding principal components of the underlying data, ordered by their eigenvalues.

The remainder of this paper is organized as follows. Section II reviews related work in online, robust, and kernel PCA. Section III formulates the problem of dynamic subspace tracking in RKHS. Section IV presents the rotation-based update mechanism in detail. Section V analyzes its convergence behavior, and Section VI presents experimental results. Section VII discusses limitations and directions for future work, and Section VIII concludes. 

\section{Background and Related Work}\label{sec:related}

\subsection{Online PCA}
The standard way to calculate the PCA of a dataset is to consider all available data at once, in a batch calculation with time complexity $\mathcal{O}(nD²)$ and space complexity $\mathcal{O}(D²)$, where $n$ is the number of samples and $D$ is the dimension of each data sample \cite{li2018}. This quadratic complexity makes batch PCA infeasible for very large datasets, or for datasets that arrive in a streaming fashion. To overcome this limitation, iterative methods have been developed which process one sample at a time and update an estimate of the principal subspace with each new sample \cite{oja1985}. Such methods, including the Generalized Hebbian Algorithm \cite{sanger1989} and the Sequential Karhunen-Loeve (SKL) transform \cite{levey2000}, each achieve per-sample time and space complexity of $\mathcal{O}(D)$. A limitation of these approaches is that they assume the underlying data distribution is static; to track a distribution that changes over time, a "forgetting factor" $\alpha < 1$ has been proposed to reduce the weight of previously observed data \cite{ross2008}. However, none of these online PCA methods are kernelized or robust to outliers.

\subsection{Kernel PCA}
Traditional PCA relies on the assumption that the data lies in a linear subspace. Kernel PCA extends this by mapping the input data into a higher-dimensional feature space via a nonlinear function $\Phi$, and performing linear PCA on the mapped features \cite{scholkopf1997}. Rather than explicitly computing $\Phi$, the "kernel trick" allows all necessary computations to be expressed in terms of a kernel function $k(\cdot,\cdot)$, which acts as a similarity measure between data samples \cite{tonin2024}. Because the resulting feature space is often infinite-dimensional, it is common in practice to use an empirical kernel map, in which a finite basis U of data samples is used to represent any new sample as a finite-dimensional vector of kernel products \cite{vong2018}. Standard kernel PCA methods are neither online nor robust, since they require calculating the kernel function between every pair of data samples, resulting in a matrix with $n^2$ entries that must be recomputed as new samples arrive.

\subsection{Robust PCA}
Because every sample in a dataset influences the calculation of a PCA, the result is necessarily sensitive to outliers. Robust PCA refers to a family of techniques that modify traditional PCA to reduce this sensitivity. One approach, known as Principal Component Pursuit \cite{candes2011}, separates the data matrix into a low-rank component and a sparse outlier component, solved via nuclear-norm minimization. A different approach applies an influence function to each sample, which measures and bounds the possible effect of a single outlier on the estimated principal components \cite{hampel2005}. Because the boundedness of the influence function can be achieved by crafting a well-chosen, differentiable robust loss function, this approach is more readily adaptable to an online, kernelized setting than approaches based on the $\ell_1$ norm, which are difficult to kernelize.

\subsection{Combining Online, Robust, and Kernel PCA}
Several works combine two of these three capabilities. Online robust PCA methods, such as Recursive Projected Compressive Sensing \cite{qiu2014} and its variants, recursively estimate a low-rank subspace and a sparse outlier component from streaming data, but do not extract nonlinear features. Online kernel PCA methods, such as the Kernel Hebbian Algorithm \cite{kim2005} and kernelized versions of the SKL transform \cite{chin2007}, extend online PCA into RKHS, but are not robust to outliers. The closest prior work to this paper is the iterative robust kernel PCA algorithm of Huang and Yeh \cite{huang2011}, which applies a robust loss function directly within the kernel PCA optimization. It can be proven, under mild assumptions, that this algorithm converges to the same subspace that would be found using batch kernel PCA on the entire dataset. However, this convergence guarantee explicitly assumes that the true underlying subspace is fixed: the algorithm is designed to converge to a static solution, and once initial convergence is achieved, its gradient-based update rule loses much of its effectiveness at tracking further changes in the underlying subspace. This is because, once the projection of a new sample onto the current subspace estimate is small, there is little residual signal left to drive further adjustment, even if the true underlying subspace has since shifted. This limitation motivates the need for a distinct update mechanism, capable of continuing to adjust the subspace estimate even after initial convergence.

\subsection{Grassmannian Geometry}\label{grassmanngeometry}
The subspace $\Gamma$ estimated by a kernel PCA algorithm can be represented as an orthonormal basis $Q$ of $d$ vectors in an M-dimensional space, where M is the dimension of the (possibly empirical) kernel feature space. Such a subspace is a point on the Grassmann manifold, the set of all $d$-dimensional subspaces of an $M$-dimensional vector space, whose geometric structure (including its geodesics and tangent spaces) has been extensively studied in the context of optimization algorithms on matrix manifolds \cite{absil2008}. The distance between two such subspaces, represented by orthonormal bases $Q_1$ and $Q_2$, can be measured by first computing the singular value decomposition of $Q_1^T Q_2$ . The resulting singular values are the cosines of the principal angles $\theta_i$ between the two subspaces, and the Grassmann distance is calculated as the root-sum-square of these principal angles.

Prior work has explored subspace tracking directly on the Grassmann manifold in an online setting, using gradient-based updates to incrementally estimate a subspace from streaming, incomplete data \cite{balzano2010}. This geometric framing suggests a natural class of update mechanisms for online subspace tracking: rather than adjusting the subspace estimate only via the gradient of a loss function, one can instead move the estimate along a path on the Grassmann manifold, toward the direction indicated by each new incoming sample. This paper adopts a related geometric perspective, but instead uses rotation matrices as the mechanism for moving the estimated subspace along such a path.

\subsection{Summary of Gaps}
Table \ref{tab1} summarizes the capabilities of the algorithms discussed above, organized along four dimensions: whether the algorithm is designed to track a dynamically changing subspace, whether it operates in an online, per-sample fashion, whether it is kernelized to extract nonlinear features, and whether it is robust to outliers. As shown, several works combine two or three of these capabilities. The closest prior work, Huang and Yeh \cite{huang2011}, combines online, kernel, and robust capabilities, but is not dynamic: it is designed to converge to a fixed subspace, and its update mechanism is not well suited to continued tracking of a subspace that changes gradually over time. This is the specific gap that this paper addresses, by introducing a rotation-based update mechanism, grounded in the Grassmannian geometry described in Section \ref{grassmanngeometry}, that enables continued dynamic tracking of the estimated subspace even after initial convergence has been achieved.

\begin{table*}[t]
\caption{Algorithm Comparison}
\begin{center}
\begin{tabular}{|l|l|c|c|c|c|}
\hline
\textbf{Algorithm} & \textbf{Reference} & \textbf{Dynamic} & \textbf{Online} & \textbf{Kernel} & \textbf{Robust} \\
\hline
Generalized Hebbian Algorithm & Sanger \cite{sanger1989} & $\times$ & \checkmark & $\times$ & $\times$ \\
\hline
SKL with forgetting factor & Ross et al. \cite{ross2008} & \checkmark & \checkmark & $\times$ & $\times$ \\
\hline
Principal Component Pursuit & Cand\`es et al. \cite{candes2011} & $\times$ & $\times$ & $\times$ & \checkmark \\
\hline
Kernel Hebbian Algorithm & Kim et al. \cite{kim2005} & $\times$ & \checkmark & \checkmark & $\times$ \\
\hline
Kernel SKL & Chin \& Suter \cite{chin2007} & $\times$ & \checkmark & \checkmark & $\times$ \\
\hline
Grassmannian Rank-One Update & Balzano et al. \cite{balzano2010} & \checkmark & \checkmark & $\times$ & $\times$ \\
\hline
Iterative Robust Kernel PCA & Huang \& Yeh \cite{huang2011} & $\times$ & \checkmark & \checkmark & \checkmark \\
\hline
\textbf{Rotation-Based Update (this paper)} & --- & \textbf{\checkmark} & \textbf{\checkmark} & \textbf{\checkmark} & \textbf{\checkmark} \\
\hline
\end{tabular}
\label{tab1}
\end{center}
\end{table*}

\section{Problem Formulation}\label{sec:formulation}

\subsection{Kernelizing Streaming Data}

Consider a stream of data samples $\mathbf{x}_t \in \mathbb{R}^D$ ($t = 1, 2, \ldots$), drawn from some underlying distribution that may change over time. To extract nonlinear features, each sample is mapped into a Hilbert space $\mathcal{H}$ via a nonlinear function $\Phi$ associated with a kernel function $k(\cdot,\cdot)$. Because $\Phi(\mathbf{x}_t)$ may be infinite-dimensional, we use a reduced-basis empirical kernel map: a finite subset $\mathbf{U} = \{\mathbf{u}_1, \ldots, \mathbf{u}_M\} \subset \mathcal{X}$ of the sample space is chosen, and each incoming sample is converted into an $M$-dimensional feature vector

\begin{equation}
\mathbf{h}_t = \Phi(\mathbf{x}_t) = \left[ k(\mathbf{u}_1, \mathbf{x}_t), \ldots, k(\mathbf{u}_M, \mathbf{x}_t) \right]^T
\label{eq:kernel_map}
\end{equation}

These $M$-dimensional vectors are treated as elements of a finite-dimensional approximation of $\mathcal{H}$, and all subsequent steps of the algorithm operate on them.

\subsection{Subspace Representation and Projection}

We represent the estimated subspace at time $t$ by an $M \times d$ matrix $\mathbf{\Gamma}_t$, whose $d$ orthonormal columns span the current estimate, together with an estimated mean $\boldsymbol{\mu}_t \in \mathbb{R}^M$. Each new feature vector $\mathbf{h}_t$ is decomposed into a component along the subspace and a component orthogonal to it. The loadings along the subspace are

\begin{equation}
\mathbf{c}_t = (\mathbf{h}_t - \boldsymbol{\mu}_t)^T \mathbf{\Gamma}_t
\label{eq:loadings}
\end{equation}

and the (half) squared distance from the centered feature vector to the subspace is

\begin{equation}
z(\mathbf{h}_t, \boldsymbol{\mu}_t, \mathbf{\Gamma}_t) = \frac{1}{2} \left\| \left( \mathbf{I} - \mathbf{\Gamma}_t \mathbf{\Gamma}_t^T \right) (\mathbf{h}_t - \boldsymbol{\mu}_t) \right\|^2_{\mathcal{H}}
\label{eq:z_distance}
\end{equation}

This distance $z$ serves both as a measure of how well the current subspace explains the new sample, and (as described next) as an input to a robustness mechanism.

\subsection{Robust Optimization Objective}

Conventional kernel PCA seeks the subspace that minimizes the total squared distance of all samples to the subspace:

\begin{equation}
\boldsymbol{\mu}, \mathbf{\Gamma} = \arg\min_{\mathbf{\Gamma}^T \mathbf{\Gamma} = \mathbf{I}} \sum_t z(\mathbf{h}_t, \boldsymbol{\mu}, \mathbf{\Gamma})
\label{eq:conventional_objective}
\end{equation}

To reduce the influence of outliers, we instead minimize a monotonically increasing, concave, differentiable robust loss function $\Psi$ applied to $z$:

\begin{equation}
\boldsymbol{\mu}, \mathbf{\Gamma} = \arg\min_{\mathbf{\Gamma}^T \mathbf{\Gamma} = \mathbf{I}} \sum_t \Psi\left(z(\mathbf{h}_t, \boldsymbol{\mu}, \mathbf{\Gamma})\right)
\label{eq:robust_objective}
\end{equation}

The derivative $\dot{\Psi}(z)$, which serves as an influence function, acts as a per-sample weight: samples farther from the current subspace estimate are assigned a smaller weight, bounding the influence any single outlier can have on the subspace update. Specific choices for $\Psi$ satisfying these properties are discussed in \cite{higuchi1998}.

\subsection{Limitation of Gradient-Based Updates}

Solving \eqref{eq:robust_objective} via gradient descent, using Lagrange multipliers to enforce the orthonormality constraint $\mathbf{\Gamma}^T \mathbf{\Gamma} = \mathbf{I}$, yields an update rule of the form

\begin{equation}
\mathbf{\Gamma}_{t+1} = \mathbf{\Gamma}_t - \eta_t \, \mathbf{g}_t^{(\Gamma)}
\label{eq:gradient_update}
\end{equation}

where $\mathbf{g}_t^{(\Gamma)}$ is the gradient of the Lagrangian with respect to $\mathbf{\Gamma}_t$, and $\eta_t$ is a learning rate. This gradient-based update is provably convergent to the batch solution when the true underlying subspace is fixed \cite{huang2011}. However, the magnitude of $\mathbf{g}_t^{(\Gamma)}$ depends on the residual component of $(\mathbf{h}_t - \boldsymbol{\mu}_t)$ orthogonal to $\mathbf{\Gamma}_t$. Once $\mathbf{\Gamma}_t$ has converged so that this residual is small for typical samples, the gradient itself becomes small, even if the true underlying subspace has since shifted, and even though new samples continue to carry information about the direction of that shift. This motivates a complementary update mechanism, described in Section~\ref{sec:rotation}, that continues to move the subspace estimate based on the \textit{direction} of new samples rather than relying solely on the magnitude of the residual gradient.

\section{Rotation-Based Subspace Update}
\label{sec:rotation}

\subsection{Motivation}

As discussed in Section~\ref{sec:formulation}, gradient-based updates to the subspace estimate $\mathbf{\Gamma}_t$ become ineffective at tracking further changes once the residual component of incoming samples orthogonal to $\mathbf{\Gamma}_t$ becomes small. We address this limitation by introducing a geometrically motivated update mechanism: rather than adjusting $\mathbf{\Gamma}_t$ only in proportion to this residual, we \textit{rotate} the entire subspace estimate toward the direction of each new incoming feature vector. Because $\mathbf{\Gamma}_t$ is constrained to have orthonormal columns, such a rotation naturally preserves this constraint without requiring an additional projection step, and can be applied even when the residual signal used by gradient-based methods is small.

\subsection{Rotation Matrix Construction}

Given two vectors $\mathbf{v}, \mathbf{w} \in \mathbb{R}^M$, we construct a rotation matrix that rotates the space spanned by $\mathbf{v}$ and $\mathbf{w}$ through a fraction $\alpha$ of the angle between them, while leaving the orthogonal complement of that plane unchanged. First, the angle $\theta$ between $\mathbf{v}$ and $\mathbf{w}$, and the desired rotation angle $\varphi$, are given by

\begin{equation}
\theta = \arccos\left( \frac{\mathbf{v}^T \mathbf{w}}{\|\mathbf{v}\| \, \|\mathbf{w}\|} \right), \qquad \varphi = \alpha \, \theta
\label{eq:rotation_angle}
\end{equation}

A $2 \times 2$ elementary rotation matrix through angle $\varphi$ is embedded into an $M \times M$ block matrix $\mathbf{B}_\varphi$ that leaves the remaining $M-2$ dimensions unchanged:

\begin{equation}
\mathbf{R}_\varphi = \begin{bmatrix} \cos\varphi & -\sin\varphi \\ \sin\varphi & \cos\varphi \end{bmatrix}, \qquad
\mathbf{B}_\varphi = \begin{bmatrix} \mathbf{R}_\varphi & \mathbf{0} \\ \mathbf{0} & \mathbf{I}_{M-2} \end{bmatrix}
\label{eq:elementary_rotation}
\end{equation}

To rotate specifically within the plane spanned by $\mathbf{v}$ and $\mathbf{w}$, we compute an orthogonal change-of-basis matrix $\mathbf{Q}$ via QR factorization of a matrix whose first two columns are $\mathbf{v}$ and $\mathbf{w}$, and whose remaining columns are chosen arbitrarily to complete a basis. The final rotation matrix is then

\begin{equation}
\mathbf{Q}, \_ = \mathrm{QR}\big( [\, \mathbf{v} \;\; \mathbf{w} \;\; \cdots \,] \big), \qquad
\mathbf{A} = \mathbf{Q} \, \mathbf{B}_\varphi \, \mathbf{Q}^T
\label{eq:rotation_matrix}
\end{equation}

The matrix $\mathbf{A}$ rotates the entire $M$-dimensional space, in the plane spanned by $\mathbf{v}, \mathbf{w}$, by an angle that is a specified fraction $\alpha$ of the angle between them, while leaving the orthogonal complement of that plane fixed.

\subsection{Rotation Type I: Projection-Based Rotation}

In the first rotation strategy, we set $\mathbf{w} = \mathbf{h}_t - \boldsymbol{\mu}_t$ and let $\mathbf{v}$ be the projection of $\mathbf{w}$ onto the current subspace estimate, $\mathbf{v} = \mathbf{\Gamma}_t \mathbf{\Gamma}_t^T (\mathbf{h}_t - \boldsymbol{\mu}_t)$. The rotation matrix $\mathbf{A}$ constructed from $\mathbf{v}$ and $\mathbf{w}$ as in \eqref{eq:rotation_matrix} is then applied to update the subspace:

\begin{equation}
\mathbf{\Gamma}_t \leftarrow \mathbf{A} \, \mathbf{\Gamma}_t
\label{eq:rotation_type1}
\end{equation}

This strategy efficiently rotates the overall subspace to align with the direction of incoming data, and is particularly effective during the initial convergence phase. However, once $\mathbf{\Gamma}_t$ is well aligned with the data, the projection $\mathbf{v}$ becomes nearly identical to $\mathbf{w}$, and the rotation \eqref{eq:rotation_type1} loses its effect. Moreover, Type~I rotation does not, by itself, encourage the individual columns of $\mathbf{\Gamma}_t$ to align with the principal directions of the data in order of decreasing eigenvalue.

\subsection{Rotation Type II: Nearest-Basis-Vector Rotation}\label{ssec:typeII}

To address this limitation, we introduce a second rotation strategy. Rather than rotating from the projection of $\mathbf{w}$ onto $\mathbf{\Gamma}_t$, we instead identify the single column $\boldsymbol{\gamma}_i$ of $\mathbf{\Gamma}_t$ that is closest to $\mathbf{w}$ in terms of cosine similarity, and rotate from that column toward $\mathbf{w}$:

\begin{equation}
\mathbf{v} = \boldsymbol{\gamma}_{i^*}, \qquad i^* = \arg\max_i \; \left| \frac{\mathbf{w}^T \boldsymbol{\gamma}_i}{\|\boldsymbol{\gamma}_i\|} \right|
\label{eq:rotation_type2_v}
\end{equation}

The resulting rotation matrix $\mathbf{A}$, constructed as in \eqref{eq:rotation_matrix} using this $\mathbf{v}$ and $\mathbf{w} = \mathbf{h}_t - \boldsymbol{\mu}_t$, is applied identically to \eqref{eq:rotation_type1}. Because $\mathbf{v}$ is a single basis vector rather than the full projection, this rotation continues to have an effect even after the subspace as a whole has converged, and it encourages individual columns of $\mathbf{\Gamma}_t$ to progressively align with the corresponding principal directions of the underlying data. In practice, we apply both Type~I and Type~II rotations at each time step, in sequence, combining the fast initial alignment of Type~I with the continued refinement enabled by Type~II.

\subsection{Robust Modulation of Rotation Magnitude}

To ensure that outliers do not disproportionately affect the subspace estimate, the rotation fraction $\alpha$ in \eqref{eq:rotation_angle} is scaled at each time step by the same influence-function weight used in the gradient-based update:

\begin{equation}
\alpha_t = \alpha_0 \, \dot{\Psi}\!\left(z(\mathbf{h}_t, \boldsymbol{\mu}_t, \mathbf{\Gamma}_t)\right)
\label{eq:robust_rotation}
\end{equation}

where $\alpha_0$ is a base rotation factor and $\dot{\Psi}(\cdot)$ is the derivative of the robust loss function introduced in Section~\ref{sec:formulation}. Samples that lie far from the current subspace estimate (and are therefore more likely to be outliers) are assigned a smaller effective rotation fraction $\alpha_t$, bounding their influence on the subspace update in the same manner as for the gradient-based update.

\subsection{Computationally Optimized Rotations}\label{ssec:computational}

A direct implementation of \eqref{eq:rotation_matrix}--\eqref{eq:rotation_type1} requires explicitly forming the $M \times M$ matrix $\mathbf{A}$, which becomes computationally infeasible for large $M$, for example, when $M$ corresponds to the flattened dimension of high-resolution video frames. We avoid this by never explicitly constructing $\mathbf{A}$ as a dense $M \times M$ matrix. Instead, the rotation is decomposed into its action on the two-dimensional subspace spanned by $\mathbf{v}$ and $\mathbf{w}$, and applied directly to the $M \times d$ matrix $\mathbf{\Gamma}_t$ through a sequence of lower-dimensional operations, so that memory and computation scale with $M \times d$ rather than $M^2$.

\subsection{Learning Rate}

Both the base rotation factor $\alpha_0$ and the gradient-based learning rate $\eta_t$ introduced in Section~\ref{sec:formulation} control a trade-off between convergence speed and stability. We keep both of these fixed over time, rather than decaying them as $t \to \infty$, as is common in classical stochastic approximation schemes. This is a deliberate design choice: a decaying learning rate would cause the algorithm to become unresponsive to further changes in the underlying subspace, which is contrary to the dynamic-tracking objective of this work. Algorithm~\ref{alg:rotation_pca} summarizes the complete per-sample update procedure described in this section.

\begin{algorithm}[hbtp]
\caption{Rotation-Based Online Robust Kernel PCA}
\label{alg:rotation_pca}
\begin{algorithmic}[1]
\Require Kernel basis $\mathbf{U}$, subspace dimension $d$, learning rate $\eta$, base rotation factor $\alpha_0$, robust loss function $\Psi$
\State Initialize $\boldsymbol{\mu}_0 \leftarrow \mathbf{0}$, $\mathbf{\Gamma}_0 \leftarrow$ orthonormal basis (e.g., first $d$ standard basis vectors)
\For{$t = 1, 2, \ldots$}
    \State Receive new sample $\mathbf{x}_t$
    \State $\mathbf{h}_t \leftarrow \Phi(\mathbf{x}_t)$ \Comment{Kernel map via Eq.~\eqref{eq:kernel_map}}
    \State $z_t \leftarrow z(\mathbf{h}_t, \boldsymbol{\mu}_{t-1}, \mathbf{\Gamma}_{t-1})$ \Comment{Distance to subspace, Eq.~\eqref{eq:z_distance}}
    \State $w_t \leftarrow \dot{\Psi}(z_t)$ \Comment{Robust influence weight}
    \State \textbf{// Gradient-based update}
    \State Compute $\mathbf{g}_t^{(\mu)}$, $\mathbf{g}_t^{(\Gamma)}$ using weight $w_t$
    \State $\boldsymbol{\mu}_t \leftarrow \boldsymbol{\mu}_{t-1} - \eta \, \mathbf{g}_t^{(\mu)}$
    \State $\mathbf{\Gamma}_t \leftarrow \mathbf{\Gamma}_{t-1} - \eta \, \mathbf{g}_t^{(\Gamma)}$
    \State $\mathbf{\Gamma}_t \leftarrow \mathrm{orth}(\mathbf{\Gamma}_t)$ \Comment{Re-orthonormalize columns}
    \State \textbf{// Rotation-based update}
    \State $\mathbf{w} \leftarrow \mathbf{h}_t - \boldsymbol{\mu}_t$
    \State $\alpha_t \leftarrow \alpha_0 \, w_t$ \Comment{Robust rotation factor, Eq.~\eqref{eq:robust_rotation}}
    \State \textit{Type I:} $\mathbf{v} \leftarrow \mathbf{\Gamma}_t \mathbf{\Gamma}_t^T \mathbf{w}$
    \State $\mathbf{A} \leftarrow$ RotationMatrix$(\mathbf{v}, \mathbf{w}, \alpha_t)$ \Comment{Eq.~\eqref{eq:rotation_matrix}}
    \State $\mathbf{\Gamma}_t \leftarrow \mathbf{A} \, \mathbf{\Gamma}_t$
    \State \textit{Type II:} $i^* \leftarrow \arg\max_i \left| \mathbf{w}^T \boldsymbol{\gamma}_i / \|\boldsymbol{\gamma}_i\| \right|$
    \State $\mathbf{v} \leftarrow \boldsymbol{\gamma}_{i^*}$
    \State $\mathbf{A} \leftarrow$ RotationMatrix$(\mathbf{v}, \mathbf{w}, \alpha_t)$
    \State $\mathbf{\Gamma}_t \leftarrow \mathbf{A} \, \mathbf{\Gamma}_t$
    \State \textbf{// Output for current sample}
    \State $\mathbf{c}_t \leftarrow (\mathbf{h}_t - \boldsymbol{\mu}_t)^T \mathbf{\Gamma}_t$ \Comment{Loadings, Eq.~\eqref{eq:loadings}}
\EndFor
\end{algorithmic}
\end{algorithm}

\section{Experimental Results}
\label{sec:results}

\subsection{Experimental Setup}

To characterize the convergence behavior of the rotation-based update, we generate synthetic streaming data from a multivariate normal distribution in a $D$-dimensional space, with mean $\boldsymbol{\mu}$ and covariance matrix $\mathbf{\Sigma} = \mathbf{Q} \mathbf{\Delta} \mathbf{Q}^T$, where the orthonormal columns of $\mathbf{Q}$ represent the true underlying directions of the data, and the eigenvalues $\lambda_j$ on the diagonal of $\mathbf{\Delta}$ represent the variance along each corresponding direction:

\begin{equation}
p(\mathbf{x} \mid \boldsymbol{\mu}, \mathbf{\Sigma}) = \frac{1}{\sqrt{(2\pi)^D |\mathbf{\Sigma}|}} \exp\left( -\frac{(\mathbf{x}-\boldsymbol{\mu})^T \mathbf{\Sigma}^{-1} (\mathbf{x}-\boldsymbol{\mu})}{2} \right)
\label{eq:mvn}
\end{equation}

Because $\mathbf{Q}$ and $\mathbf{\Delta}$ are known by construction, this synthetic setting allows the estimated subspace $\mathbf{\Gamma}_t$ to be directly compared against the ground truth at every time step. We report two performance metrics: the variance-weighted Grassmann distance between the estimated and true subspace, and the cosine similarity between individual estimated eigenvectors and their ground-truth counterparts.

\subsection{Effect of Learning Rate}

We generate data in a $D=100$-dimensional space with a subspace dimension of $d=10$, with additional Gaussian noise added in all dimensions, and run Algorithm~\ref{alg:rotation_pca} multiple times, each with a different learning rate $\eta$. Fig.~\ref{fig:learning_rate}(a) shows the variance-weighted subspace distance as a function of the number of samples processed, for several values of $\eta$; Fig.~\ref{fig:learning_rate}(b) shows the corresponding cosine similarity of the first estimated eigenvector. As $\eta$ increases, the subspace distance converges faster, at the expense of a slightly noisier estimate of the individual eigenvectors. This confirms that $\eta$ provides the expected trade-off between convergence speed and stability of the estimate.

\subsection{Rotation Improves Convergence}

\begin{figure*}[!t]
  \centering
  
  \begin{minipage}{0.98\textwidth}
    \centering
    \begin{subfigure}[htbp]{0.45\textwidth} 
        \includegraphics[width=\textwidth]{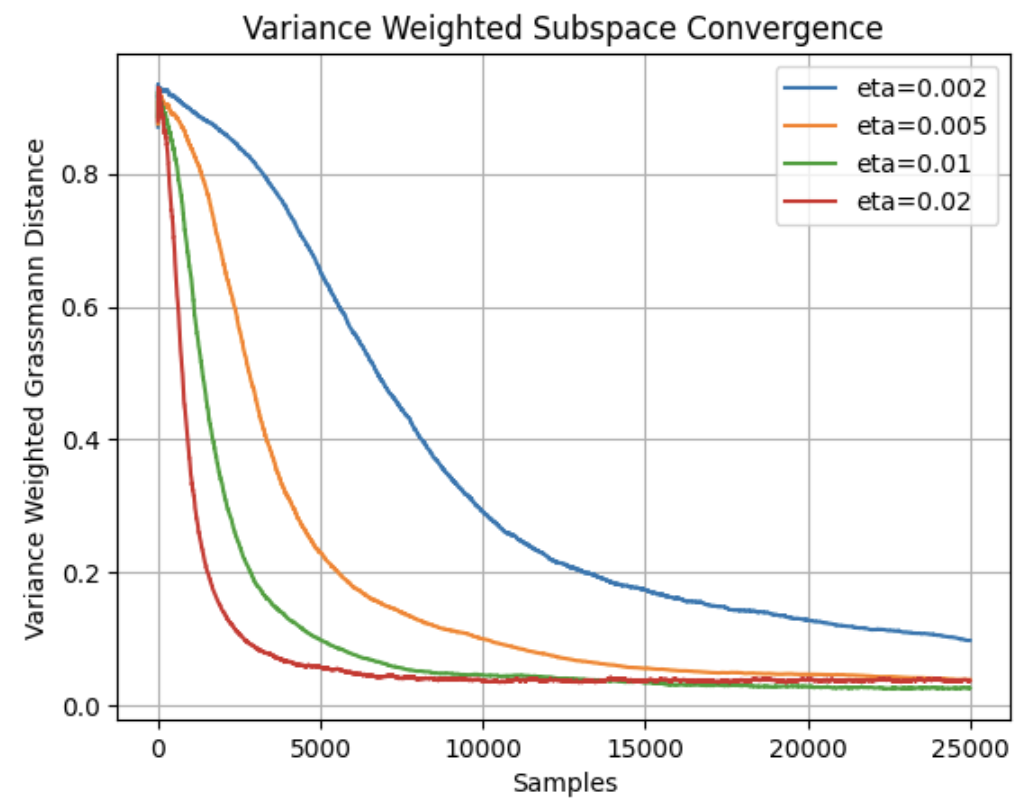} 
    \end{subfigure}\qquad 
    \begin{subfigure}[htbp]{0.45\textwidth} 
        \includegraphics[width=\textwidth]{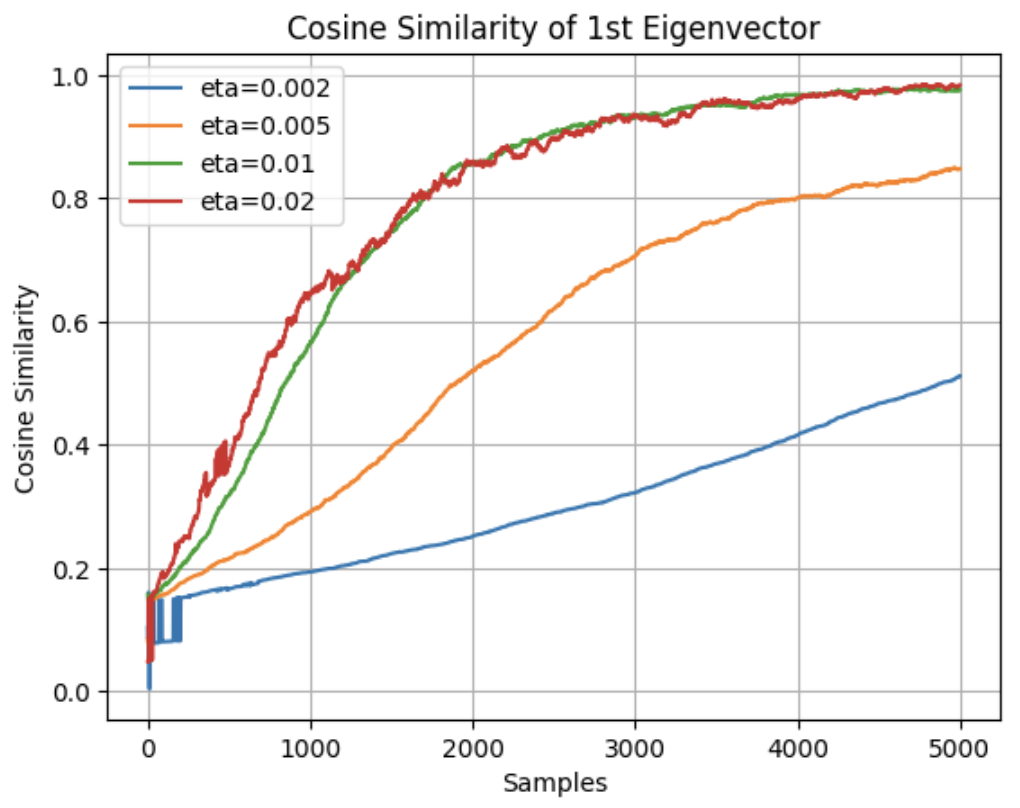} 
    \end{subfigure}
    \caption{(a) Variance-weighted subspace distance and (b) cosine similarity of the first eigenvector, for increasing learning rate $\eta$. Larger $\eta$ yields faster convergence at the cost of increased noise.}
    \label{fig:learning_rate}
  \end{minipage}
  
  \vspace{2em} 
  
  \begin{minipage}{0.48\textwidth}
    \centering
    \includegraphics[width=0.99\textwidth]{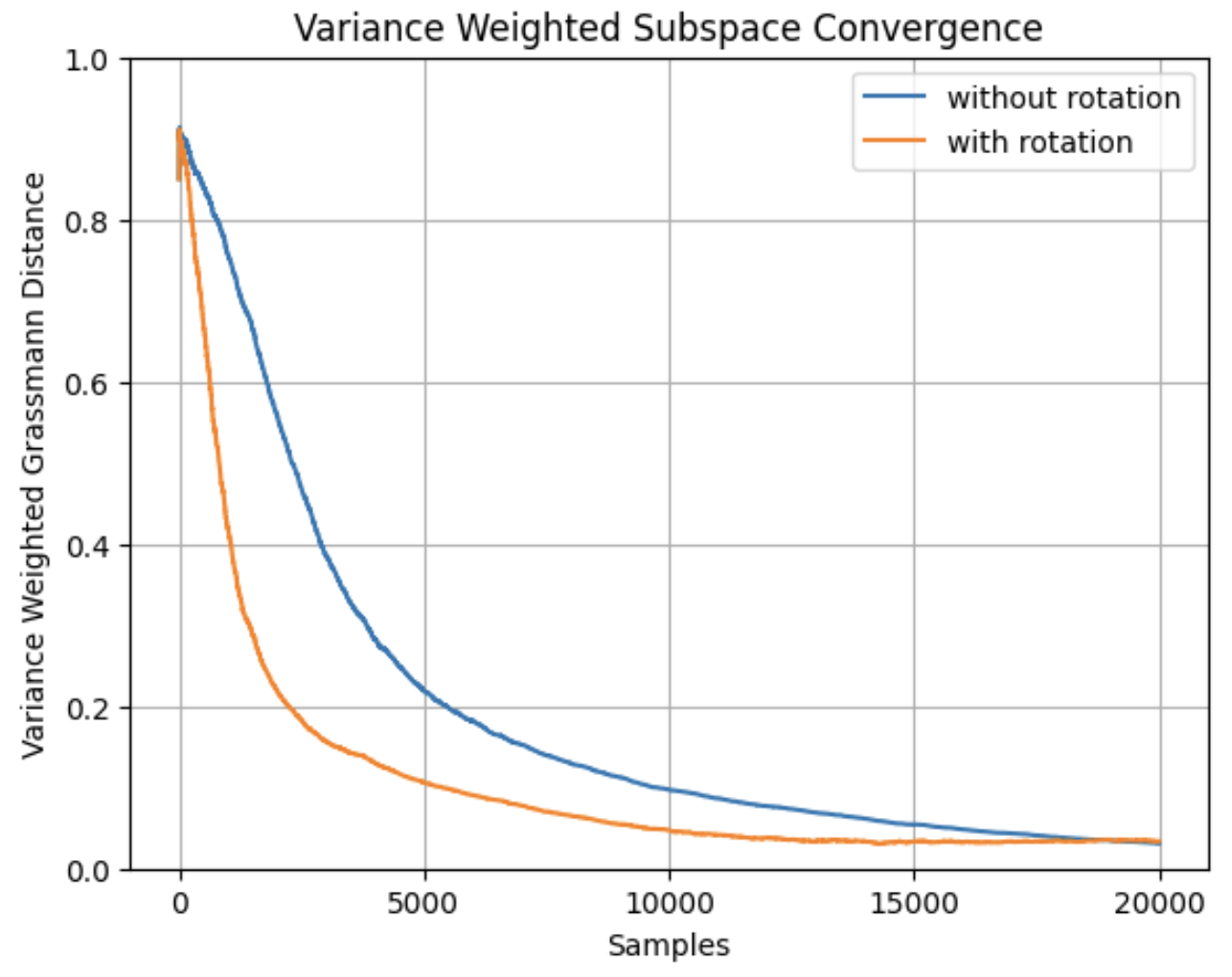}
    \caption{Variance-weighted subspace distance versus number of samples, with and without the rotation-based update enabled. Enabling rotation yields substantially faster convergence.}
    \label{fig:rotation_ablation}
  \end{minipage}
  \hfill
  \begin{minipage}{0.48\textwidth}
    \centering
    \includegraphics[width=\textwidth]{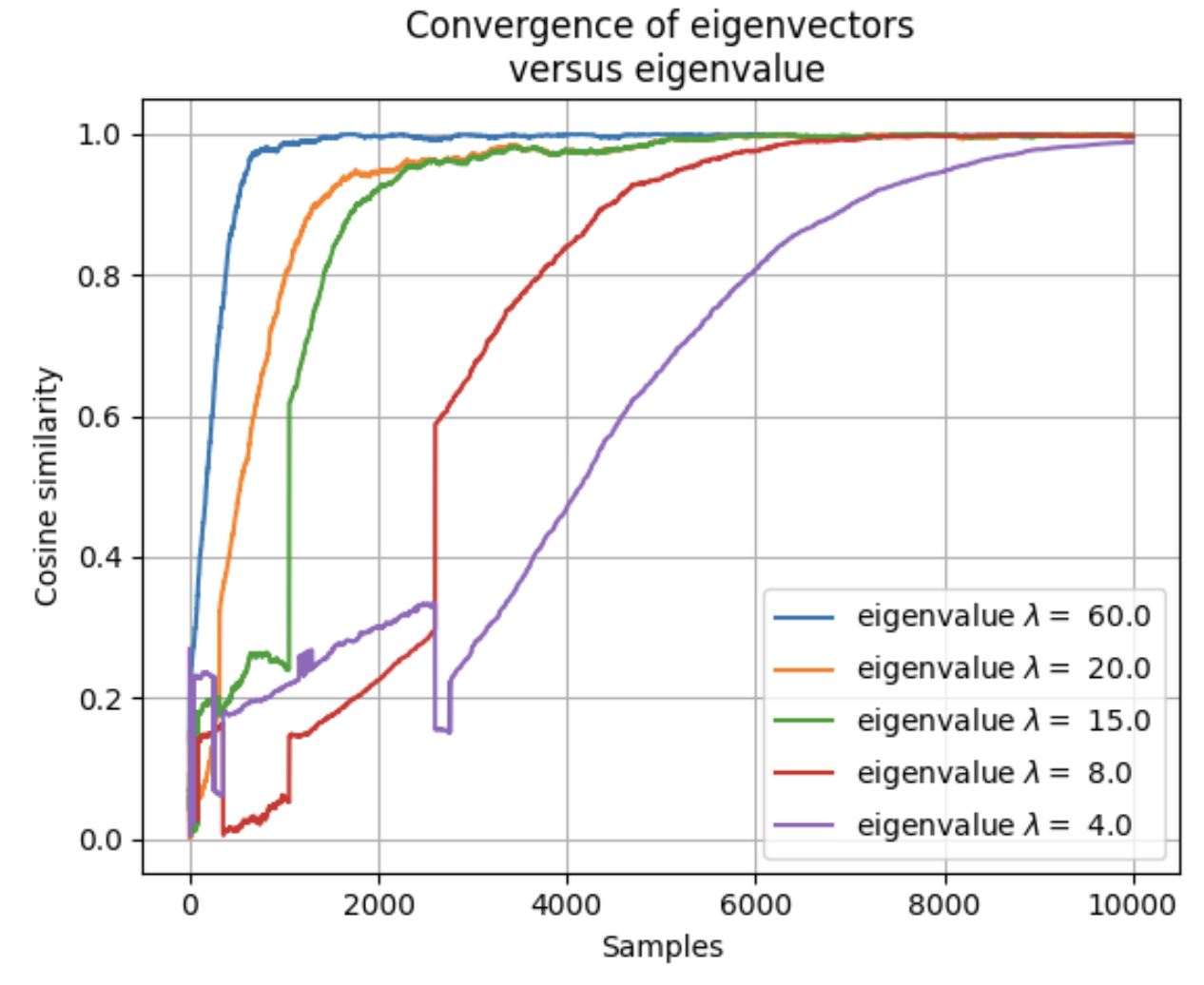}
    \caption{Cosine similarity of five estimated eigenvectors to ground truth, versus number of samples, for eigenvalues $\lambda \in \{60, 20, 16, 8, 4\}$. Eigenvectors with larger eigenvalues converge first.}
    \label{fig:eigenvalue_order}
  \end{minipage}
\end{figure*}

To isolate the contribution of the rotation-based update, we generate data under the same conditions ($D=100$, $d=10$, with Gaussian noise), and run Algorithm~\ref{alg:rotation_pca} twice: once with the rotation steps (lines 12--19) disabled, so that the subspace is updated using only the gradient-based mechanism of Section~\ref{sec:formulation}, and once with the full algorithm, including both Type~I and Type~II rotations. In both cases, all other hyperparameters, including the learning rate $\eta$, are held fixed. Fig.~\ref{fig:rotation_ablation} shows the variance-weighted subspace distance as a function of the number of samples processed, with and without rotation enabled. The subspace distance to ground truth converges substantially faster when the rotation-based update is enabled, confirming that rotation provides a convergence benefit beyond what is achievable through gradient descent alone.

\subsection{Convergence Order by Eigenvalue}

To examine how convergence depends on the magnitude of the underlying eigenvalues, we generate data with $D=100$ and $d=10$, setting the eigenvalues along the first five basis vectors to $\lambda \in \{60, 20, 16, 8, 4\}$. Fig.~\ref{fig:eigenvalue_order} shows the cosine similarity between each of the five estimated eigenvectors and their corresponding ground-truth directions, as a function of the number of samples processed. The eigenvector associated with the largest eigenvalue converges most quickly, and each subsequently smaller eigenvalue takes correspondingly longer to converge, with the eigenvector of smallest eigenvalue converging last. This ordering is consistent with the geometric intuition underlying Type~II rotation (Section~\ref{ssec:typeII}): directions with larger eigenvalues account for a larger share of the variance in the data, so a larger fraction of incoming samples carry a component along those directions, driving faster rotation-based alignment.

\section{Discussion}
\label{sec:discussion}

The experimental results in Section~\ref{sec:results} demonstrate that the rotation-based update mechanism converges empirically in a manner consistent with its intended design: faster convergence is achieved when rotation is enabled compared to gradient descent alone, the learning rate hyperparameter trades off convergence speed against estimate stability in an interpretable way, and convergence proceeds in an order that reflects the relative importance of each principal direction. Together, these results support the central claim of this paper: that per-sample rotations in Reproducing Kernel Hilbert Space provide an effective mechanism for dynamic subspace tracking, beyond what is achievable through gradient-based updates alone.

\subsection{Scope of Experimental Validation}

The experiments in this paper use synthetic data with known ground truth, in order to isolate and rigorously evaluate the convergence benefit of the rotation-based update mechanism itself. This choice allows the estimated subspace to be compared directly against a known reference at every time step, without the confounding effects of labeling noise or downstream task variance that would be introduced by real-world data. Validating this mechanism within a complete online, robust, kernel PCA pipeline on real-world application data (such as streaming network traffic or video images) is the subject of ongoing work.

\subsection{Theoretical Limitations}

We do not provide a formal convergence theorem for the rotation-based update on the Grassmann manifold. The empirical results in Section~\ref{sec:results} are consistent with convergence under the tested conditions, but they do not establish a convergence bound, nor characterize the trajectory of $\mathbf{\Gamma}_t$ on the manifold in the way that has been done for gradient-based Grassmannian subspace tracking~\cite{balzano2010}. Establishing such a theoretical guarantee for the rotation-based update (for example, by characterizing the rotation step as a discrete approximation of a geodesic path on the Grassmann manifold) is a natural and important direction for future work.

\subsection{Practical Considerations}

The rotation-based update introduces two additional hyperparameters beyond those required for the gradient-based update alone: the base rotation factor $\alpha_0$, and the choice of applying Type~I and/or Type~II rotation at each step. While we found a fixed, non-decaying value of $\alpha_0$ to be effective across the synthetic settings tested, we have not conducted a systematic study of how this hyperparameter should be tuned for datasets with different noise characteristics or different rates of subspace drift. In addition, the computational cost of the rotation-based update, while designed in Section~\ref{ssec:computational} to scale favorably with the feature dimension $M$, still requires constructing a QR factorization at each time step. A more detailed computational comparison against the cost of the gradient-based update alone is left for future work.

\section{Conclusion}
\label{sec:conclusion}

This paper introduced a rotation-based update mechanism for dynamic subspace tracking in online kernel PCA. Unlike gradient-based updates, which lose effectiveness once the residual component of incoming samples orthogonal to the estimated subspace becomes small, rotation-based updates continue to move the subspace estimate based on the direction of new samples, enabling continued tracking of a subspace that changes over time. We introduced two complementary rotation strategies: one that efficiently aligns the overall subspace during initial convergence, and one that continues to refine the ordering of individual basis vectors according to their corresponding eigenvalues. We also showed how the magnitude of each rotation can be modulated by a robust influence function to bound the effect of outliers. Through controlled experiments on synthetic streaming data with known ground truth, we demonstrated that the rotation-based update converges substantially faster than gradient descent alone, and that this convergence behavior is consistent with the underlying geometric intuition of the mechanism. While a formal convergence theorem on the Grassmann manifold remains an open problem, the empirical results presented here establish rotation-based updates as an effective and efficient mechanism for dynamic subspace tracking, and lay the groundwork for validating this mechanism on a complete online, robust, kernel PCA pipeline on real-world streaming data.

\end{document}